\documentclass[sigconf]{acmart}
\usepackage{booktabs} 

\usepackage[english]{babel}
\usepackage{moresize}
\usepackage{amsmath}
\usepackage{algorithmic}
\usepackage{balance}
\usepackage{comment}
\usepackage{paralist}
\usepackage{bm}
\usepackage{pgfplots}

\usepackage{flushend}
\usepackage[english]{babel}
\usepackage[latin1]{inputenc}
\usepackage{mathrsfs}
\usepackage{graphicx}

\usepackage{amssymb}
\usepackage{amsfonts}
\usepackage{url}
\usepackage{longtable}
\usepackage{rotating}
\usepackage{multirow}
\usepackage{mathrsfs}
\usepackage{subfigure}
\usepackage{enumitem}
\usepackage[linesnumbered,algoruled,boxed,lined]{algorithm2e}
\usepackage{adjustbox}
\usepackage{hyperref}
\usepackage{pgfplots}
\usepackage{filecontents}
\usepackage{mydef}

\definecolor{tblue}{RGB}{31,119,180}
\definecolor{torange}{RGB}{255,127,14}
\definecolor{tgreen}{RGB}{44,160,44}
\definecolor{tred}{RGB}{214,39,40}
\definecolor{tpurple}{RGB}{148,103,189}

\newcommand{\hide}[1]{} 

\newcommand{\ie}{\textit{i}.\textit{e}.}
\newcommand{\eg}{\textit{e}.\textit{g}.}

\copyrightyear{2024}
\acmYear{2024}
\setcopyright{acmlicensed}\acmConference[KDD '24]{Proceedings of the 30th ACM SIGKDD Conference on Knowledge Discovery and Data Mining}{August 25--29, 2024}{Barcelona, Spain}
\acmBooktitle{Proceedings of the 30th ACM SIGKDD Conference on Knowledge Discovery and Data Mining (KDD '24), August 25--29, 2024, Barcelona, Spain}
\acmDOI{10.1145/3637528.3671460}
\acmISBN{979-8-4007-0490-1/24/08}

\begin{document}



\title{A Survey of Large Language Models for Graphs}

\author{Xubin Ren}
\affiliation{%
  \institution{University of Hong Kong}
  \city{Hong Kong}
  \country{China}}
\email{xubinrencs@gmail.com}

\author{Jiabin Tang}
\affiliation{%
  \institution{University of Hong Kong}
  \city{Hong Kong}
  \country{China}}
\email{jiabintang77@gmail.com}

\author{Dawei Yin}
\affiliation{%
  \institution{Baidu Inc.}
  \city{Beijing}
  \country{China}}
\email{yindawei@acm.org}

\author{Nitesh Chawla}
\affiliation{
  \institution{University of Notre Dame}
  \city{Indiana}
  \country{USA}}
\email{nchawla@nd.edu}

\author{Chao Huang}
\authornote{Corresponding author}
\affiliation{%
  \institution{University of Hong Kong}
  \city{Hong Kong}
  \country{China}}
\email{chaohuang75@gmail.com}

\renewcommand{\shorttitle}{A Survey of Large Language Models for Graphs}
\renewcommand{\shortauthors}{Xubin Ren, Jiabin Tang, Dawei Yin, Nitesh Chawla, \& Chao Huang}

\begin{abstract}
Graphs are an essential data structure utilized to represent relationships in real-world scenarios. Prior research has established that Graph Neural Networks (GNNs) deliver impressive outcomes in graph-centric tasks, such as link prediction and node classification. Despite these advancements, challenges like data sparsity and limited generalization capabilities continue to persist. Recently, Large Language Models (LLMs) have gained attention in natural language processing. They excel in language comprehension and summarization. Integrating LLMs with graph learning techniques has attracted interest as a way to enhance performance in graph learning tasks. In this survey, we conduct an in-depth review of the latest state-of-the-art LLMs applied in graph learning and introduce a novel taxonomy to categorize existing methods based on their framework design. We detail four unique designs: i) GNNs as Prefix, ii) LLMs as Prefix, iii) LLMs-Graphs Integration, and iv) LLMs-Only, highlighting key methodologies within each category. We explore the strengths and limitations of each framework, and emphasize potential avenues for future research, including overcoming current integration challenges between LLMs and graph learning techniques, and venturing into new application areas. This survey aims to serve as a valuable resource for researchers and practitioners eager to leverage large language models in graph learning, and to inspire continued progress in this dynamic field. We consistently maintain the related open-source materials at \color{blue}{\url{https://github.com/HKUDS/Awesome-LLM4Graph-Papers}}.
\end{abstract}

\maketitle

\section{Introduction}
\label{sec:introduction}

Graphs, comprising nodes and edges that signify relationships, are essential for illustrating real-world connections across various domains. These include social networks~\cite{myers2014information, li2024urbangpt}, molecular graphs~\cite{InstructMol}, recommender systems~\cite{he2020lightgcn, DCCF}, and academic networks~\cite{hu2020open}. This structured data form is integral in mapping complex interconnections relevant to a wide range of applications.

In recent years, Graph Neural Networks (GNNs)~\cite{wu2020comprehensive} have emerged as a powerful tool for a variety of tasks, including node classification~\cite{GIN} and link prediction~\cite{zhang2018link}. By passing and aggregating information across nodes and iteratively refining node features through supervised learning, GNNs have achieved remarkable results in capturing structural nuances and enhancing model accuracy. To accomplish this, GNNs leverage graph labels to guide the learning process. Several notable models have been proposed in the literature, each with its own strengths and contributions. For instance, Graph Convolutional Networks (GCNs)~\cite{gcn} have been shown to be effective in propagating embeddings across nodes, while Graph Attention Networks (GATs)~\cite{gat} leverage attention mechanisms to perform precise aggregation of node features. Additionally, Graph Transformers~\cite{GTN, GFormer} employ self-attention and positional encoding to capture global signals among the graph, further improving the expressiveness of GNNs. To address scalability challenges in large graphs, methods such as Nodeformer~\cite{NodeFormer} and DIFFormer~\cite{DIFFormer} have been proposed. These approaches employ efficient attention mechanisms and differentiable pooling techniques to reduce computational complexity while maintaining high levels of accuracy. Despite these advancements, current GNN methodologies still face several challenges. For example, data sparsity remains a significant issue, particularly in scenarios where the graph structure is incomplete or noisy~\cite{you2021graph}. Moreover, the generalization ability of GNNs to new graphs or unseen nodes remains an open research question, with recent works highlighting the need for more robust and adaptive models~\cite{garg2020generalization, OpenGraph, zhao2024graphany}.

Large Language Models (LLMs)~\cite{zhao2023survey}, which show great generalization abilities for unseen tasks~\cite{BERT, T5, wang2023far}, have emerged as powerful tools in various research fields, including natural language processing~\cite{achiam2023gpt}, computer vision~\cite{liu2024visual, liu2024improved}, and information retrieval~\cite{zhu2023collaborative, hou2024large, lin2024data}.
The advent of LLMs has sparked significant interest within the graph learning community~\cite{huang2024large, jin2023large, li2023survey}, prompting investigations into the potential of LLMs to enhance performance on graph-related tasks. Researchers have explored various approaches to leverage the strengths of LLMs for graph learning, resulting in a new wave of methods that combine the power of LLMs with graph neural networks.
One promising direction is to develop prompts that enable LLMs to understand graph structures and respond to queries effectively. For instance, approaches such as InstructGLM~\cite{InstructGLM} and NLGraph~\cite{NLGraph} have designed specialized prompts that allow LLMs to reason over graph data and generate accurate responses.
Alternatively, other methods have integrated GNNs to feed tokens into the LLMs, allowing them to understand graph structures more directly. For example, GraphGPT~\cite{GraphGPT} and GraphLLM~\cite{GraphLLM} use GNNs to encode graph data into tokens, which are then fed into the LLMs for further processing. This synergy between LLMs and GNNs has not only improved task performance but also demonstrated impressive zero-shot generalization capabilities, where the models can accurately answer queries about unseen graphs or nodes.

In this survey, we offer a systematic review of the advancements in Large Language Models (LLMs) for graph applications, and we explore potential avenues for future research. Unlike prior surveys that categorize studies based on the role of LLMs~\cite{jin2023large, li2023survey} or focus primarily on integrating LLMs with knowledge graphs~\cite{pan2024unifying}, our work highlights the model framework design, particularly the inference and training processes, to distinguish between existing taxonomies. This perspective allows readers to gain a deeper understanding of how LLMs effectively address graph-related challenges. We identify and discuss four distinct architectural approaches: i) \textit{GNNs as Prefix}, ii) \textit{LLMs as Prefix}, iii) \textit{LLMs-Graphs Integration}, and iv) \textit{LLMs-Only}, each illustrated with representative examples. In summary, the contributions of our work can be summarized as:
\begin{itemize}[leftmargin=*]
    \item \textbf{Comprehensive Review of LLMs for Graph Learning.} We offer a comprehensive review of the current state-of-the-art Large Language Models (LLMs) for graph learning, elucidating their strengths and pinpointing their limitations.
    \item \textbf{Novel Taxonomy for Categorizing Research.} We introduce a novel taxonomy for categorizing existing research based on their framework design, which provides a deeper insight into how LLMs can be seamlessly integrated with graph learning.
    \item \textbf{Future Research Avenues.} We also explore potential avenues for future research, including addressing the prevalent challenges in merging LLMs with graph learning methods and venturing into novel application areas.
\end{itemize}
\tikzstyle{redleaf}=[draw=edgered,
    rounded corners,minimum height=1em,
    fill=contentred!40,text opacity=1, align=center,
    fill opacity=.5,  text=black,align=left,font=\scriptsize,
    inner xsep=3pt,
    inner ysep=1pt,
]

\tikzstyle{greenleaf}=[draw=edgegreen,
    rounded corners,minimum height=1em,
    fill=contentgreen!40,text opacity=1, align=center,
    fill opacity=.5,  text=black,align=left,font=\scriptsize,
    inner xsep=3pt,
    inner ysep=1pt,
]

\tikzstyle{blueleaf}=[draw=edgeblue,
    rounded corners,minimum height=1em,
    fill=contentblue!40,text opacity=1, align=center,
    fill opacity=.5,  text=black,align=left,font=\scriptsize,
    inner xsep=3pt,
    inner ysep=1pt,
]

\tikzstyle{yellowleaf}=[draw=edgeyellow,
    rounded corners,minimum height=1em,
    fill=contentyellow!40,text opacity=1, align=center,
    fill opacity=.5,  text=black,align=left,font=\scriptsize,
    inner xsep=3pt,
    inner ysep=1pt,
]

\tikzstyle{blackmiddle}=[draw=black,
    rounded corners,minimum height=1em,
    fill=output-white!40,text opacity=1, align=center,
    fill opacity=.5,  text=black,align=left,font=\scriptsize,
    inner xsep=3pt,
    inner ysep=1pt,
]

\tikzstyle{redmiddle}=[draw=edgered,
    rounded corners,minimum height=1em,
    fill=output-white!40,text opacity=1, align=center,
    fill opacity=.5,  text=black,align=left,font=\scriptsize,
    inner xsep=3pt,
    inner ysep=1pt,
]

\tikzstyle{greenmiddle}=[draw=edgegreen,
    rounded corners,minimum height=1em,
    fill=output-white!40,text opacity=1, align=center,
    fill opacity=.5,  text=black,align=left,font=\scriptsize,
    inner xsep=3pt,
    inner ysep=1pt,
]

\tikzstyle{bluemiddle}=[draw=edgeblue,
    rounded corners,minimum height=1em,
    fill=output-white!40,text opacity=1, align=center,
    fill opacity=.5,  text=black,align=left,font=\scriptsize,
    inner xsep=3pt,
    inner ysep=1pt,
]

\tikzstyle{yellowmiddle}=[draw=edgeyellow,
    rounded corners,minimum height=1em,
    fill=output-white!40,text opacity=1, align=center,
    fill opacity=.5,  text=black,align=left,font=\scriptsize,
    inner xsep=3pt,
    inner ysep=1pt,
]

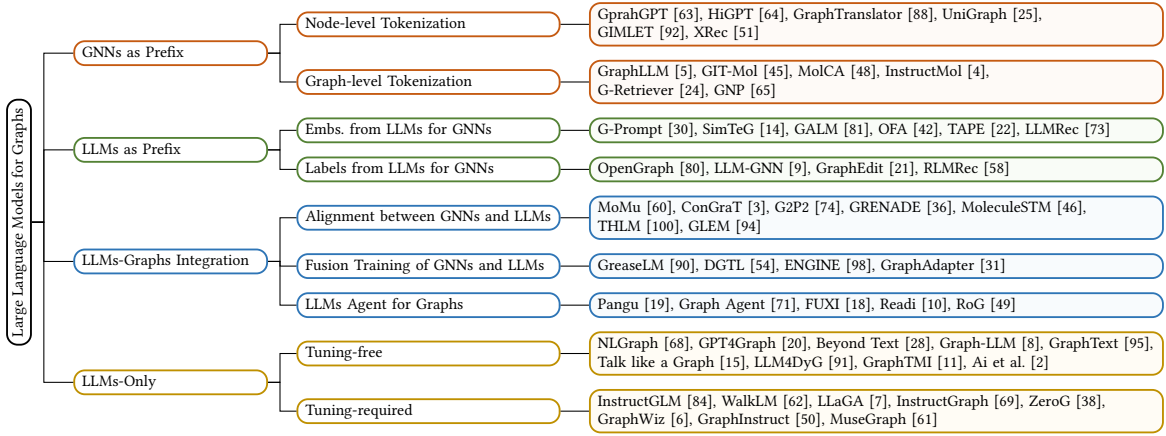
\begin{figure*}[ht]
\centering
\begin{forest}
  for tree={
  forked edges,
  grow=east,
  reversed=true,
  anchor=base west,
  parent anchor=east,
  child anchor=west,
  base=middle,
  font=\scriptsize,
  rectangle,
  line width=0.7pt,
  draw=output-black,
  rounded corners,align=left,
  minimum width=2em,
    s sep=5pt,
    inner xsep=3pt,
    inner ysep=1pt,
  },
  where level=1{text width=4.5em}{},
  where level=2{text width=6em,font=\scriptsize}{},
  where level=3{font=\scriptsize}{},
  where level=4{font=\scriptsize}{},
  where level=5{font=\scriptsize}{},
  [Large Language Models for Graphs, blackmiddle, rotate=90, anchor=north, edge=output-black
    [GNNs as Prefix, redmiddle, edge=output-black,text width=7.4em
        [Node-level Tokenization, redmiddle, text width=10.3em, edge=output-black
            [GprahGPT~\cite{GraphGPT}{,} HiGPT~\cite{HiGPT}{,} GraphTranslator~\citep{GraphTranslator}{,} UniGraph~\cite{UniGraph}{,} \\GIMLET~\cite{GIMLET}{,} XRec~\cite{XRec}, redleaf, text width=23.3em, edge=output-black]
        ]
        [Graph-level Tokenization, redmiddle, text width=10.3em, edge=output-black
            [GraphLLM~\cite{GraphLLM}{,} GIT-Mol~\cite{GIT-Mol}{,} MolCA~\cite{MolCA}{,} InstructMol~\cite{InstructMol}{,} \\G-Retriever~\cite{G-Retriever}{,} GNP~\cite{GNP}, redleaf, text width=23.3em, edge=output-black]
        ]
    ]
    [LLMs as Prefix, greenmiddle, edge=output-black, text width=7.4em
        [Embs. from LLMs for GNNs, greenmiddle, text width=10.3em, edge=output-black
            [G-Prompt~\cite{G-Prompt}{,} SimTeG~\cite{SimTeG}{,} GALM~\cite{GALM}{,} OFA~\cite{OFA}{,} TAPE~\cite{TAPE}{,} LLMRec~\cite{LLMRec}, greenleaf, text width=23.3em, edge=output-black]
        ]
        [Labels from LLMs for GNNs, greenmiddle, text width=10.3em, edge=output-black
            [OpenGraph~\cite{OpenGraph}{,} LLM-GNN~\cite{LLM-GNN}{,} GraphEdit~\cite{GraphEdit}{,} RLMRec~\cite{RLMRec}, greenleaf, text width=23.3em, edge=output-black]
        ]
    ]
    [LLMs-Graphs Integration, bluemiddle, edge=output-black, text width=7.4em
        [Alignment between GNNs and LLMs, bluemiddle, text width=10.3em, edge=output-black
            [MoMu~\cite{MoMu}{,} ConGraT~\cite{ConGraT}{,} G2P2~\cite{G2P2}{,} GRENADE~\cite{GRENADE}{,} MoleculeSTM~\cite{MoleculeSTM}{,} \\THLM~\cite{THLM}{,} GLEM~\cite{GLEM}, blueleaf, text width=23.3em, edge=output-black]
        ]
        [Fusion Training of GNNs and LLMs, bluemiddle, text width=10.3em, edge=output-black
            [GreaseLM~\cite{GreaseLM}{,} DGTL~\cite{DGTL}{,} ENGINE~\cite{ENGINE}{,} GraphAdapter~\cite{GraphAdapter}, blueleaf, text width= 23.3em, edge=output-black]
        ]
        [LLMs Agent for Graphs, bluemiddle, text width=10.3em, edge=output-black
            [Pangu~\cite{Pangu}{,} Graph Agent~\cite{GraphAgent}{,} FUXI~\cite{FUXI}{,} Readi~\cite{Readi}{,} RoG~\cite{RoG}, blueleaf, text width= 23.3em, edge=output-black]
        ]
    ]
    [LLMs-Only, yellowmiddle, edge=output-black, text width=7.4em
        [Tuning-free, yellowmiddle, edge=output-black, text width=10.3em
            [NLGraph~\cite{NLGraph}{,} GPT4Graph~\cite{GPT4Graph}{,} Beyond Text~\cite{BeyondText}{,} Graph-LLM~\cite{Graph-LLM}{,} GraphText~\cite{GraphText}{,}\\ Talk like a Graph~\cite{Talklikeagraph}{,} LLM4DyG~\cite{LLM4DyG}{,} GraphTMI~\cite{GraphTMI}{,} Ai et al.~\cite{GraphMM}, yellowleaf, text width=23.3em, edge=output-black]
        ]
        [Tuning-required, yellowmiddle, edge=output-black, text width=10.3em
            [InstructGLM~\cite{InstructGLM}{,} WalkLM~\cite{WalkLM}{,} LLaGA~\cite{LLaGA}{,} InstructGraph~\cite{InstructGraph}{,} ZeroG~\cite{ZeroG}{,}\\ GraphWiz~\cite{GraphWiz}{,} GraphInstruct~\cite{GraphInstruct}{,} MuseGraph~\cite{MuseGraph},  yellowleaf, text width=23.3em, edge=output-black]
        ]
     ]
  ]
\end{forest}
\caption{The proposed taxonomy of Large Language Models (LLMs) for graphs, featuring representative works.}
\label{fig:taxonomy_of_pGMs}
\end{figure*}

\section{Preliminaries and Taxonomy}
\label{sec:taxonomy}

In this section, we first provide essential background knowledge on large language models and graph learning. Then, we present our taxonomy of large language models for graphs.

\subsection{Definitions}
\noindent \textbf{Graph-Structured Data}. In computer science, a graph $\mathcal{G}=(\mathcal{V},\mathcal{E})$ is a non-linear data structure that consists of a set of nodes $\mathcal{V}$, and a set of edges $\mathcal{E}$ connecting these nodes. Each edge $e \in \mathcal{E}$ is associated with a pair of nodes $(u, v)$, where $u$ and $v$ are the endpoints of the edge. The edge may be directed, meaning it has a orientation from $u$ to $v$, or undirected, meaning it has no orientation. Furthermore, A Text-Attributed Graph (TAG) is a graph that assigns a sequential text feature (\ie, sentence) to each node, denoted as $\textbf{t}_v$, which is widely used in the era of large language models. The text-attributed graph can be formally represented as $\mathcal{G}_S=(\mathcal{V},\mathcal{E},\mathcal{T})$, where $\mathcal{T}$ is the set of text features.

\noindent \textbf{Graph Neural Networks (GNNs)} are deep learning architectures for graph-structured data that aggregate information from neighboring nodes to update node embeddings. Formally, the update of a node embedding $\mathbf{h}_{v} \in \mathbb{R}^{d}$ in each GNN layer can be represented as: 
\begin{align}
    \mathbf{h}_{v}^{(l+1)} = \psi(\phi(\{ \mathbf{h}_{v'}^{(l)}: v' \in \mathcal{N}(v) \}), \mathbf{h}_{v}^{(l)}), 
\end{align}
where $v' \in \mathcal{N}(v)$ denotes a neighbor node of $v$, and $\phi(\cdot)$ and $\psi(\cdot)$ are aggregation and update functions, respectively. By stacking $L$ GNN layers, the final node embeddings can be used for downstream graph-related tasks such as node classification and link prediction.

\noindent \textbf{Large Language Models (LLMs)}. Language Models (LMs) is a statistical model that estimate the probability distribution of words for a given sentence. Recent research has shown that LMs with billions of parameters exhibit superior performance in solving a wide range of natural language tasks (\eg, translation, summarization and instruction following), making them Large Language Models (LLMs). In general, most recent LLMs are built with transformer blocks that use a query-key-value (QKV)-based attention mechanism to aggregate information in the sequence of tokens. Based on the direction of attention, LLMs can be categorized into two types (given a sequence of tokens $\mathbf{x} = [x_0, x_1, ..., x_n]$):
\begin{itemize}[leftmargin=*]
    \item \textit{\textbf{Masked Language Modeling (MLM)}}. Masked Language Modeling is a popular pre-training objective for LLMs that involves masking out certain tokens in a sequence and training the model to predict the masked tokens based on the surrounding context. Specifically, the model takes into account both the left and right context of the masked token to make accurate predictions:
    \begin{align}
        p(x_i | x_0, x_1, ..., x_n).
    \end{align}
    Representative models include BERT~\cite{BERT} and RoBERTa~\cite{RoBERTa}. 
    \item \textit{\textbf{Causal Language Modeling (CLM)}}. Causal Language Modeling is another popular training objective for LLMs that involves predicting the next token in a sequence based on the previous tokens. Specifically, the model only considers the left context of the current token to make accurate predictions:
    \begin{align}
        p(x_i | x_0, x_1, ..., x_{i-1})
    \end{align}
    Notable examples include the GPT (\eg, ChatGPT) and Llama~\cite{Llama}.
\end{itemize}

\subsection{Taxonomy}
In this survey, we present our taxonomy focusing on the model inference pipeline that processes both graph data and text with LLMs. Specifically, we summarize four main types of model architecture design for large language models for graphs, as follows:
\begin{itemize}[leftmargin=*]
    \item \textit{\textbf{GNNs as Prefix}}. GNNs serve as the first component to process graph data and provide structure-aware tokens (e.g., node-level, edge-level, or graph-level tokens) for LLMs for inference.
    \item \textit{\textbf{LLMs as Prefix}}. LLMs first process graph data with textual information and then provide node embeddings or generated labels for improved training of graph neural networks.
    \item \textit{\textbf{LLMs-Graphs Integration}}. In this line, LLMs achieve a higher level of integration with graph data, such as fusion training or alignment with GNNs, and also build LLM-based agents to interact with graph information.
    \item \textit{\textbf{LLMs-Only}}. This line designs practical prompting methods to ground graph-structured data into sequences of words for LLMs to infer, while some also incorporate multi-modal tokens.
\end{itemize}

\section{Large Language Models for Graphs}
\label{sec:llmsgraph}

\subsection{GNNs as Prefix}

In this section, we discuss the application of graph neural networks (GNNs) as structural encoders to enhance the understanding of graph structures by LLMs, thereby benefiting various downstream tasks, i.e., \emph{GNNs as Prefix}. In these methods, GNNs generally play the role of a tokenizer, encoding graph data into a graph token sequence rich in structural information, which is then input into LLMs to align with natural language. These methods can generally be divided into two categories: i) \textit{Node-level Tokenization}: each node of the graph structure is input into the LLM, aiming to make the LLM understand fine-grained node-level structural information and distinguish relationships. ii) \textit{Graph-level Tokenization}: the graph is compressed into a fixed-length token sequence using a specific pooling method, aiming to capture high-level global semantic information of the graph structure.

\subsubsection{\bf Node-level Tokenization}
For some downstream tasks in graph learning, such as node classification and link prediction, the model needs to model the fine-grained structural information at node level, and distinguish the semantic differences between different nodes. Traditional GNNs usually encode a unique representation for each node based on the information of neighboring nodes, and directly perform downstream node classification or link prediction. In this line, the node-level tokenization method is utilized, which can retain the unique structural representation of each node as much as possible, thereby benefiting downstream tasks.

Within this line, \textbf{GraphGPT}~\cite{GraphGPT} proposes to initially align the graph encoder with natural language semantics through text-graph grounding, and then combine the trained graph encoder with the LLM using a projector. Through the two-stage instruction tuning paradigm, the model can directly complete various graph learning downstream tasks with natural language, thus perform strong zero-shot transferability and multi-task compatibility. The proposed Chain-of-Thought distillation method empowers GraphGPT to migrate to complex tasks with small parameter sizes. Then, \textbf{HiGPT}~\cite{HiGPT} proposes to combine the language-enhanced in-context heterogeneous graph tokenizer with LLMs, solving the challenge of relation type heterogeneity shift between different heterogeneous graphs. Meanwhile, the two-stage heterogeneous graph instruction-tuning injects both homogeneity and heterogeneity awareness into the LLM. And the Mixture-of-Thought (MoT) method combined with various prompt engineering further solves the common data scarcity problem in heterogeneous graph learning. \textbf{GIMLET}~\cite{GIMLET}, as a unified graph-text model, leverages natural language instructions to address the label insufficiency challenge in molecule-related tasks, effectively alleviating the reliance on expensive lab experiments for data annotation. It employs a generalized position embedding and attention mechanism to encode both graph structures and textual instructions as a unified token combination that is fed into a transformer decoder. 
\textbf{GraphTranslator}~\cite{GraphTranslator} proposes the use of a translator with shared self-attention to align both the target node and instruction, and employs cross attention to map the node representation encoded by the graph model to fixed-length semantic tokens. The proposed daul-phase training paradigm empowers the LLM to make predictions based on language instructions, providing a unified solution for both pre-defined and open-ended graph-based tasks. 
Instead of using pre-computed node features of varying dimensions, \textbf{UniGraph}~\cite{UniGraph} leverages Text-Attributed Graphs for unifying node representations, featuring a cascaded architecture of language models and graph neural networks as backbone networks. In recent research on recommendation systems, \textbf{XRec}~\cite{XRec} has been proposed as a method that utilizes the encoded user/item embeddings from graph neural networks as collaborative signals. These signals are then integrated into each layer of large language models, enabling the generation of explanations for recommendations, even in zero-shot scenarios.

\begin{figure}[t]
    \centering
    \includegraphics[width=0.90\columnwidth]{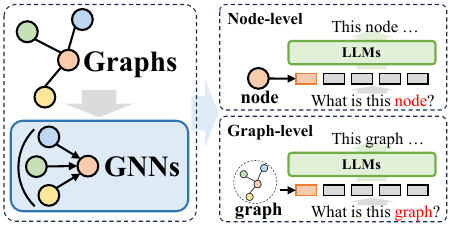}
    \vspace{-0.1in}
    \caption{GNNs as Prefix.}
    \vspace{-0.2in}
    \label{fig:gnnprefix}
\end{figure}

\subsubsection{\bf Graph-level Tokenization}
On the other hand, to adapt to other graph-level tasks, models need to be able to extract global information from node representations, to obtain high-level graph semantic tokens. In the method of \textit{GNN as Prefix}, Graph-level tokenization abstracts node representations into unified graph representations through various "pooling" operations, further enhancing various downstream tasks.

Within this domain, \textbf{GraphLLM}~\cite{GraphLLM} utilizes a graph transformer that incorporates the learnable query and positional encoding to encode the graph structure and obtain graph representations through pooling. These representations are directly used as graph-enhanced prefix for prefix tuning in the LLM, demonstrating remarkable effectiveness in fundamental graph reasoning tasks. \textbf{MolCA}~\cite{MolCA} with Cross-Modal Projector and Uni-Modal Adapter is a method that enables a language model to understand both text- and graph-based molecular contents through the proposed dual-stage pre-training and fine-tuning stage. It employs a cross-modal projector implemented as a Q-Former to connect a graph encoder's representation space and a language model's text space, and a uni-modal adapter for efficient adaptation to downstream tasks. 
\textbf{InstructMol}~\cite{InstructMol} introduces a projector that aligns the molecular graph encoded by the graph encoder with the molecule's Sequential information and natural language instructions, with the first stage of Alignment Pretraining and the second stage of Task-specific Instruction Tuning enabling the model to achieve excellent performance in various drug discovery-related molecular tasks. 
\textbf{GIT-Mol}~\cite{GIT-Mol} further unifies the graph, text, and image modalities through interaction cross-attention between different modality encoders, and aligns these three modalities, enabling the model to simultaneously perform four downstream tasks: captioning, generation, recognition, and prediction.
\textbf{GNP}~\cite{GNP} employs cross-modality pooling to integrate the node representations encoded by the graph encoder with the natural language tokens, resulting in a unified graph representation. This representation is aligned with the instruction through the LLM to demonstrate superiority in commonsense and biomedical reasoning tasks. Recently, \textbf{G-Retriever}~\cite{G-Retriever} utilizes retrieval-augmented techniques to obtain subgraph structures. It completes various downstream tasks in GraphQA (Graph Question Answering) through the collaboration of graph encoder and LLMs.

\subsubsection{\bf Discussion}
The \textit{GNN as Prefix} approach aligns the modeling capability of GNNs with the semantic modeling capability of LLMs, demonstrating unprecedented generalization, \ie, zero-shot capability, in various graph learning downstream tasks and real-world applications. However, despite the effectiveness of the aforementioned approach, the challenge lies in whether the \textit{GNN as Prefix} method remains effective for non-text-attributed graphs. Additionally, the optimal coordination between the architecture and training of GNNs and LLMs remains an unresolved question.

\subsection{LLMs as Prefix}

The methods presented in this section leverage the information produced by large language models to improve the training of graph neural networks. This information includes textual content, labels, or embeddings derived from the large language models. These techniques can be categorized into two distinct groups: i) \textit{Embeddings from LLMs for GNNs}, which involves using embeddings generated by large language models for graph neural networks, and ii) \textit{Labels from LLMs for GNNs}, which involves integrating labels generated by large language models for graph neural networks.

\subsubsection{\textbf{Embeddings from LLMs for GNNs}} The inference process of graph neural networks involves passing node embeddings through the edges and then aggregating them to obtain the next-layer node embeddings. In this process, the initial node embeddings are diverse across different domains. For instance, ID-based embeddings in recommendation systems or bag-of-words embeddings in citation networks can be unclear and non-diverse. Sometimes, the poor quality of initial node embeddings can result in suboptimal performance of GNNs. Furthermore, the lack of a universal design for node embedders makes it challenging to address the generalization capability of GNNs in unseen tasks with different node sets. Fortunately, the works in this line leverage the powerful language summarization and modeling capabilities of LLMs to generate meaningful and effective embeddings for GNNs' training.

In this domain, \textbf{G-Prompt}~\cite{G-Prompt} adds a GNN layer at the end of a pre-trained language models (PLMs) to achieve graph-aware fill-masking self-supervised learning. By doing so, G-Prompt can generate task-specific, explainable node embeddings for downstream tasks using prompt tuning. \textbf{SimTeG}~\cite{SimTeG} first leverages parameter-efficient fine-tuning on the text embeddings obtained by LLMs for downstream tasks (e.g., node classification). Then, the node embeddings are fed into GNNs for inference. Similarly, \textbf{GALM}~\cite{GALM} utilizes BERT as a pre-language model to encode text embeddings for each node. Then, the model is pre-trained through unsupervised learning tasks, such as link prediction, to minimize empirical loss and find optimal model parameters, which enables GALM to be applied for various downstream tasks. Recently, \textbf{OFA}~\cite{OFA} leverages LLMs to unify graph data from different domains into a common embedding space for cross-domain learning. It also uses LLMs to encode task-relevant text descriptions for constructing prompt graphs, allowing the model to perform specific tasks based on context. \textbf{TAPE}~\cite{TAPE} uses customized prompts to query LLMs, generating both prediction and text explanation for each node. Then, DeBERTa is fine-tuned to convert the text explanations into node embeddings for GNNs. Finally, GNNs can use a combination of the original text features, explanation features, and prediction features to predict node labels. In the field of recommendation, \textbf{LLMRec}~\cite{LLMRec} achieves graph augmentation on user-item interaction data using GPT-3.5, which not only filters out noise interactions and adds meaningful training data, but also enriches the initial node embeddings for users and items with generated rich textual profiles, ultimately improving the performance of recommenders.

\begin{figure}[t]
    \centering
    \includegraphics[width=0.90\columnwidth]{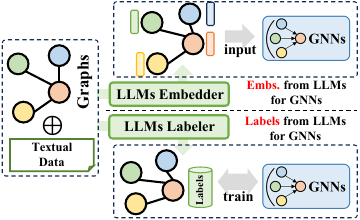}
    \vspace{-0.1in}
    \caption{LLMs as Prefix.}
    \vspace{-0.2in}
    \label{fig:llmprefix}
\end{figure}

\subsubsection{\textbf{Labels from LLMs for GNNs}} Another approach leverages the generated labels from large language models as supervision to improve the training of graph neural networks. Notably, the supervised labels in this context are not limited to categorized labels in classification tasks, but can take various forms such as embeddings, graphs, and more. The generated information from the LLMs is not used as input to the GNNs, but rather forms the supervision signals for better optimization, which enables GNNs to achieve higher performance on various graph-related tasks.

Follow this line, \textbf{OpenGraph}~\cite{OpenGraph} employs LLMs to generate nodes and edges, mitigating the issue of sparse training data. The generation process for nodes and edges is refined using the Gibbs sampling algorithm and a tree-of-prompt strategy, which is then utilized to train the graph foundation model. \textbf{LLM-GNN}~\cite{LLM-GNN} leverages LLMs as annotators to generate node category predictions with confidence scores, which serve as labels. Post-filtering is then employed to filter out low-quality annotations while maintaining label diversity. Finally, the generated labels are used to train GNNs. \textbf{GraphEdit}~\cite{GraphEdit} leverages the LLMs to build an edge predictor, which is used to evaluate and refine candidate edges against the original graph's edges. In recommender systems, \textbf{RLMRec}~\cite{RLMRec} leverages LLMs to generate text descriptions of user/item preferences. These descriptions are then encoded as semantic embeddings to guide the representation learning of ID-based recommenders using contrastive and generative learning techniques~\cite{ren2024comprehensive}.

\subsubsection{\textbf{Discussion}} Despite the progress made by the aforementioned methods in enhancing graph learning performance, a limitation persists in their decoupled nature, where LLMs are not co-trained with GNNs, resulting in a two-stage learning process. This decoupling is often due to computational resource limitations arising from the large size of the graph or the extensive parameters of LLMs. Consequently, the performance of the GNNs is heavily dependent on the pre-generated embeddings/labels of LLMs or even the design of task-specific prompts.

\subsection{LLMs-Graphs Integration}

The methods introduced in this section aim to further integrate large language models with graph data, encompassing various methodologies that enhance not only the ability of LLMs to tackle graph tasks but also the parameter learning of GNNs. These works can be categorized into three types: i) \textit{Fusion Training of GNNs and LLMs}, which aims to achieve fusion-co-training of the parameters of both models; ii) \textit{Alignment between GNNs and LLMs}, which focuses on achieving representation or task alignment between the two models; and iii) \textit{LLMs Agent for Graphs}, which builds an autonomous agent based on LLMs to plan and solve graph tasks.

\subsubsection{\textbf{Alignment between GNNs and LLMs}} In general, GNNs and LLMs are designed to handle different modalities of data, with GNNs focusing on structural data and LLMs focusing on textual data. This results in different feature spaces for the two models. To address this issue and make both modalities of data more beneficial for the learning of both GNNs and LLMs, several methods use techniques such as contrastive learning or Expectation-Maximization (EM) iterative training to align the feature spaces of the two models. This enables better modeling of both graph and text information, resulting in improved performance on various tasks.

Within this topic, \textbf{MoMu}~\cite{MoMu} is a multimodal molecular foundation model that includes two separate encoders, one for handling molecular graphs (GIN) and another for handling text data (BERT). It uses contrastive learning to pre-train the model on a dataset of molecular graph-text pairs. This approach enables MoMu to directly imagine new molecules from textual descriptions. Also in the bioinfo domain, \textbf{MoleculeSTM}~\cite{MoleculeSTM} combines the chemical structure information of molecules (\ie, molecular graph) with their textual descriptions (\ie, SMILES strings), and uses a contrastive learning to jointly learn the molecular structure and textual descriptions. It show great performance on multiple benchmark tests, including structure-text retrieval, text-based editing tasks, and molecular property prediction. Similarly, in \textbf{ConGraT}~\cite{ConGraT}, a contrastive graph-text pretraining technique is proposed to align the node embeddings encoded by LMs and GNNs simultaneously. The experiments are conducted on social networks, citation networks, and link networks, and show great performance on node and text classification as well as link prediction tasks. Furthermore, \textbf{G2P2}~\cite{G2P2, G2P2*} enhances graph-grounded contrastive pre-training by proposing three different types of alignment: text-node, text-summary, and node-summary alignment. This enables G2P2 to leverage the rich semantic relationships in the graph structure to improve text classification performance in low-resource environments. \textbf{GRENADE}~\cite{GRENADE} is a graph-centric language model that proposes graph-centric contrastive learning and knowledge alignment to achieve both node-level and neighborhood-level alignment based on the node embeddings encoded from GNNs and LMs. This enables the model to capture text semantics and graph structure information through self-supervised learning, even in the absence of human-annotated labels. In addition to contrastive learning, \textbf{THLM}~\cite{THLM} leverages BERT and HGNNs to encode node embeddings and uses a positive-negative classification task with negative sampling to improve the alignment of embeddings from two different modalities. Recently, \textbf{GLEM}~\cite{GLEM} adopts an efficient and effective solution that integrates graph structure and language learning through a variational expectation-maximization (EM) framework. By iteratively using LMs and GNNs to provide labels for each other in node classification, GLEM aligns their capabilities in graph tasks.

\begin{figure}[t]
    \centering
    \includegraphics[width=0.90\columnwidth]{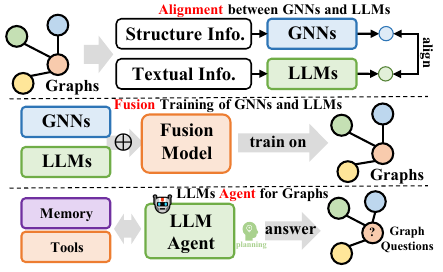}
    \vspace{-0.1in}
    \caption{LLMs-Graphs Integration.}
    \vspace{-0.2in}
    \label{fig:llmgraphs}
\end{figure}

\subsubsection{\textbf{Fusion Training of GNNs and LLMs}} Although alignment between the representations of GNNs and LLMs achieves co-optimization and embedding-level alignment of the two models, they remain separate during inference. To achieve a higher level of integration between LLMs and GNNs, several works have focused on designing a deeper fusion of the architecture of the modules, such as transformer layers in LLMs and graph neural layers in GNNs. Co-training GNNs and LLMs can result in a win-win bi-directional benefit for both modules in graph tasks.

Along this line, \textbf{GreaseLM}~\cite{GreaseLM} integrates transformer layers and GNN layers by designing a specific forward propagation layer that enables bidirectional information passing between LM and GNN through special interaction markers and interaction nodes. This approach allows language context representations to be grounded in structured world knowledge, while subtle linguistic differences (such as negation or modifiers) can affect the representation of the knowledge graph, which enables GreaseLM to achieve high performance on Question-Answering tasks. \textbf{DGTL}~\cite{DGTL} proposes disentangled graph learning to leverage GNNs to encode disentangled representations, which are then injected into each transformer layer of the LLMs. This approach enables the LLMs to be aware of the graph structure and leverage the gradient from the LLMs to fine-tune the GNNs. By doing so, DGTL achieves high performance on both citation network and e-commerce graph tasks. \textbf{ENGINE}~\cite{ENGINE} adds a lightweight and tunable G-Ladder module to each layer of the LLM, which uses a message-passing mechanism to integrate structural information. This enables the output of each LLM layer (\ie, token-level representations) to be passed to the corresponding G-Ladder, where the node representations are enhanced and then used for downstream tasks such as node classification. More directly, \textbf{GraphAdapter}~\cite{GraphAdapter} uses a fusion module (typically a multi-layer perceptrons) to combine the structural representations obtained from GNNs with the contextual hidden states of LLMs (\eg, the encoded node text). This enables the structural information from the GNN adapter to complement the textual information from the LLMs, resulting in a fused representation that can be used for supervision training and prompting for downstream tasks.

\subsubsection{\textbf{LLMs Agent for Graphs}} With the powerful capabilities of LLMs in understanding instructions and self-planning to solve tasks, an emerging research direction is to build autonomous agents based on LLMs to tackle human-given or research-related tasks. Typically, an agent consists of a memory module, a perception module, and an action module to enable a loop of observation, memory recall, and action for solving given tasks. In the graph domain, LLMs-based agents can interact directly with graph data to perform tasks such as node classification and link prediction.

In this field, \textbf{Pangu}~\cite{Pangu} pioneered the use of LMs to navigate KGs. In this approach, the agent is designed as a symbolic graph search algorithm, providing a set of potential search paths for the language models to evaluate in response to a given query. The remaining path is then utilized to retrieve the answer. \textbf{Graph Agent (GA)}~\cite{GraphAgent} converts graph data into textual descriptions and generates embedding vectors, which are stored in long-term memory. During inference, GA retrieves similar samples from long-term memory and integrates them into a structured prompt, which is used by LLMs to explain the potential reasons for node classification or edge connection. \textbf{FUXI}~\cite{FUXI} framework integrates customized tools and the ReAct~\cite{yao2022react} algorithm to enable LLMs to act as agents that can proactively interact with KGs. By leveraging tool-based navigation and exploration of data, these agents perform chained reasoning to progressively build answers and ultimately solve complex queries efficiently and accurately. \textbf{Readi}~\cite{Readi} is another approach that first uses in-context learning and chain-of-thought prompts to generate reasoning paths with multiple constraints, which are then instantiated based on the graph data. The instantiated reasoning paths are merged and used as input to LLMs to generate an answer. This method has achieved significant performance improvements on KGQA (knowledge graph question answering) and TableQA (table question answering) tasks. Recently, \textbf{RoG}~\cite{RoG} is proposed to answer graph-retaled question in three steps: planning, retrieval, and reasoning. In the planning step, it generates a set of associated paths based on the structured information of the knowledge graph according to the problem. In the retrieval step, it uses the associated paths generated in the planning stage to retrieve the corresponding reasoning paths from the KG. Finally, it uses the retrieved reasoning paths to generate the answer and explanation for the problem using LLMs.

\subsubsection{\textbf{Discussion}} 
The integration of LLMs and graphs has shown promising progress in minimizing the modality gap between structured data and textual data for solving graph-related tasks. By combining the strengths of LLMs in language understanding and the ability of graphs to capture complex relationships between entities, we can enable more accurate and flexible reasoning over graph data. However, despite the promising progress, there is still room for improvement in this area.
One of the main challenges in integrating LLMs and graphs is scalability. In alignment and fusion training, current methods often use small language models or fix the parameters of LLMs, which limits their ability to scale to larger graph datasets. Therefore, it is crucial to explore methods for scaling model training with larger models on web-scale graph data, which can enable more accurate and efficient reasoning over large-scale graphs.
Another challenge in this area is the limited interaction between graph agents and graph data. Current methods for graph agents often plan and execute only once, which may not be optimal for complex tasks requiring multiple runs. Therefore, it is necessary to investigate methods for agents to interact with graph data multiple times, refining their plans and improving their performance based on feedback from the graph. This can enable more sophisticated reasoning over graph data and improve the accuracy of downstream tasks.
Overall, the integration of LLMs and graphs is a promising research direction with significant potential for advancing the state-of-the-art in graph learning. By addressing the aforementioned challenges and developing more advanced methods for integrating LLMs and graphs, we can enable more accurate and flexible reasoning over graph data and unlock new applications in areas such as knowledge graph reasoning, molecular modeling, and social network analysis.

\subsection{LLMs-Only}

\begin{figure}[t]
    \centering
    \includegraphics[width=0.90\columnwidth]{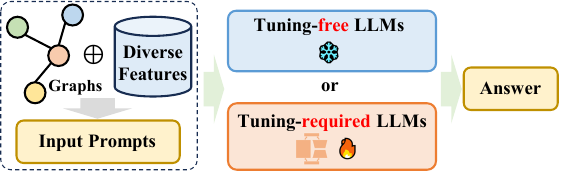}
    \vspace{-0.1in}
    \caption{LLMs-Only.}
    \vspace{-0.2in}
    \label{fig:llmonly}
\end{figure}

In this section, we will elaborate in detail on the direct application of LLMs for various graph-oriented tasks, namely the \textit{LLMs-Only} category. These methods aim to allow LLMs to directly accept graph structure information, understand it, and perform inference for various downstream tasks in combination with this information. These methods can mainly be divided into two broad categories: i) \textbf{Tuning-free} methods aim to design prompts that LLMs can understand to express graphs, directly prompting pre-trained LLMs to perform graph-oriented tasks; ii) \textbf{Tuning-required} approaches focus on converting graphs into sequences in a specific way and aligning graph token sequences and natural language token sequences using fine-tuning methods.

\subsubsection{\bf Tuning-free}
Given the unique structured characteristics of graph data, two critical challenges arise: effectively constructing a graph in natural language format and determining whether Large Language Models (LLMs) can accurately comprehend graph structures as represented linguistically. To address these issues, tuning-free approaches are being developed to model and infer graphs solely within the text space, thereby exploring the potential of pre-trained LLMs for enhanced structural understanding.

\textbf{NLGraph}~\cite{NLGraph}, \textbf{GPT4Graph}~\cite{GPT4Graph} and \textbf{Beyond Text}~\cite{BeyondText} collectively examine the capabilities of LLMs in understanding and reasoning with graph data. NLGraph proposes a benchmark for graph-based problem solving and introduces instruction-based approaches, while GPT4Graph and Beyond Text investigate the proficiency of LLMs in comprehending graph structures and emphasizes the need for advancements in their graph processing capabilities. 
And \textbf{Graph-LLM}~\cite{Graph-LLM} explores the potential of LLMs in graph machine learning, focusing on the node classification task. Two pipelines, \textit{LLMs-as-Enhancers} and \textit{LLMs-as-Predictors}, are investigated to leverage LLMs' extensive common knowledge and semantic comprehension abilities. Through comprehensive studies, it provides original observations and insights that open new possibilities for utilizing LLMs in learning on graphs. 
Meanwhile, \textbf{GraphText}~\cite{GraphText} translates graphs into natural language by deriving a graph-syntax tree and processing it with an LLM. It offers training-free graph reasoning and enables interactive graph reasoning, showcasing the unexplored potential of LLMs. 
\textbf{Talk like a Graph}~\cite{Talklikeagraph} conducts an in-depth examination of text-based graph encoder functions for LLMs, evaluating their efficacy in transforming graph data into textual format to enhance LLMs' capabilities in executing graph reasoning tasks, and proposes the GraphQA benchmark to systematically measure the influence of encoding strategies on model performance.
And \textbf{LLM4DyG}~\cite{LLM4DyG} benchmarks the spatial-temporal comprehension of LLMs on dynamic graphs, introducing tasks that evaluate both temporal and spatial understanding, and suggests the Disentangled Spatial-Temporal Thoughts (DST2) prompting technique for improved performance.
To facilitate the integration of multimodality, \textbf{GraphTMI}~\cite{GraphTMI} presents an innovative approach to integrating graph data with LLMs, introducing diverse modalities such as text, image, and motif encoding to enhance LLMs' efficiency in processing complex graph structures, and proposes the GraphTMI benchmark for evaluating LLMs in graph structure analysis, revealing that the image modality outperforms text and prior GNNs in balancing token limits and preserving essential information. \textbf{Ai et al.}~\cite{GraphMM} introduces a multimodal framework for graph understanding and reasoning, utilizing image encoding and GPT-4V's advanced capabilities to interpret and process diverse graph data, while identifying challenges in Chinese OCR and complex graph types, suggesting directions for future enhancements in AI's multimodal interaction and graph data processing.

\subsubsection{\bf Tuning-required}
Due to the limitations of expressing graph structural information using pure text, the recent mainstream approach is to align graphs as node token sequences with natural language token sequences when inputting them to LLMs. In contrast to the aforementioned \textit{GNN as Prefix} approach, the \textit{Tuning-required LLM-only} approach discards the graph encoder and adopts a specific arrangement of graph token sequences, along with carefully designed embeddings of graph tokens in prompts, achieving promising performances in various downstream graph-related tasks. 

\textbf{InstructGLM}~\cite{InstructGLM} introduces an innovative framework for graph representation learning that combines natural language instructions with graph embeddings to fine-tune LLMs. This approach allows LLMs to effectively process graph structures without relying on specialized GNN architectures. \textbf{WalkLM}~\cite{WalkLM} integrates language models with random walks to create unsupervised attributed graph embeddings, focusing on the technical innovation of transforming graph entities into textual sequences and utilizing graph-aware fine-tuning. This technique captures both attribute semantics and graph structures. Recently, \textbf{LLaGA}~\cite{LLaGA} has utilized node-level templates to restructure graph data into organized sequences, which are then mapped into the token embedding space. This allows Large Language Models to process graph-structured data with enhanced versatility, generalizability, and interpretability. \textbf{InstructGraph}~\cite{InstructGraph} proposes a methodological approach to improve LLMs for graph reasoning and generation through structured format verbalization, graph instruction tuning, and preference alignment. This aims to bridge the semantic gap between graph data and textual language models, and to mitigate the issue of hallucination in LLM outputs.

\textbf{ZeroG}~\cite{ZeroG} then leverages a language model to encode node attributes and class semantics, employing prompt-based subgraph sampling and lightweight fine-tuning strategies to address cross-dataset zero-shot transferability challenges in graph learning.
Furthermore, \textbf{GraphWiz}~\cite{GraphWiz} utilizes GraphInstruct, an instruction-tuning dataset, to augment language models for addressing various graph problems, employing Direct Preference Optimization (DPO)~\cite{rafailov2024direct} to enhance the clarity and accuracy of reasoning processes.
\textbf{GraphInstruct}~\cite{GraphInstruct} presents a comprehensive benchmark of 21 graph reasoning tasks, incorporating diverse graph generation methods and detailed reasoning steps to enhance LLMs with improved graph understanding and reasoning capabilities.
And, \textbf{MuseGraph}~\cite{MuseGraph} fuses the capabilities of LLMs with graph mining tasks through a compact graph description mechanism, diverse instruction generation, and graph-aware instruction tuning, enabling a generic approach for analyzing and processing attributed graphs.

\subsubsection{\bf Discussion}
The \textit{LLMs-Only} approach is an emerging research direction that explores the potential of pre-training Large Language Models specifically for interpreting graph data and merging graphs with natural language instructions. The main idea behind this approach is to leverage the powerful language understanding capabilities of LLMs to reason over graph data and generate accurate responses to queries. However, effectively transforming large-scale graphs into text prompts and reordering graph token sequences to preserve structural integrity without a graph encoder present significant ongoing challenges. These challenges arise due to the complex nature of graph data, which often contains intricate relationships between nodes and edges, as well as the limited ability of LLMs to capture such relationships without explicit guidance. As such, further research is needed to develop more advanced methods for integrating LLMs with graph data and overcoming the aforementioned challenges.
\section{Future Directions}
\label{sec:future directions}

In this section, we explore several open problems and potential future directions in the field of large language models for graphs.

\vspace{-0.05in}
\subsection{LLMs for Multi-modal Graphs}
Recent studies have demonstrated the remarkable ability of large language models to process and understand multi-modal data~\cite{wu2023next}, such as images~\cite{liu2024visual} and videos~\cite{zhang2023video}. This capability has opened up new avenues for integrating LLMs with multi-modal graph data, where nodes may contain features from multiple modalities~\cite{liang2024survey}. By developing multi-modal LLMs that can process such graph data, we can enable more accurate and comprehensive reasoning over graph structures, taking into account not only textual information but also visual, auditory, and other types of data.

\vspace{-0.1in}
\subsection{Efficiency and Less Computational Cost}
In the current landscape, the substantial computational expenses associated with both the training and inference phases of LLMs pose a significant limitation~\cite{ding2023parameter, gao2022parameter}, impeding their capacity to process large-scale graphs that encompass millions of nodes. This challenge is further compounded when attempting to integrate LLMs with GNNs, as the fusion of these two powerful models becomes increasingly arduous due to the aforementioned computational constraints~\cite{GLEM}. Consequently, the necessity to discover and implement efficient strategies for training LLMs and GNNs with reduced computational costs becomes paramount. This is not only to alleviate the current limitations but also to pave the way for the enhanced application of LLMs in graph-related tasks, thereby broadening their utility and impact in the field of data science.

\vspace{-0.1in}
\subsection{Tackling Different Graph Tasks}
The prevailing methodologies LLMs have primarily centered their attention on conventional graph-related tasks, such as link prediction and node classification. However, considering the remarkable capabilities of LLMs, it is both logical and promising to delve into their potential in tackling more complex and generative tasks, including but not limited to graph generation~\cite{zhu2022survey}, graph understanding, and graph-based question answering~\cite{huang2019knowledge}. By expanding the horizons of LLM-based approaches to encompass these intricate tasks, we can unlock a myriad of new opportunities for their application across diverse domains. For instance, in the realm of drug discovery, LLMs could facilitate the generation of novel molecular structures; in social network analysis, they could provide deeper insights into intricate relationship patterns; and in knowledge graph construction, they could contribute to the creation of more comprehensive and contextually accurate knowledge bases.

\subsection{User-Centric Agents on Graphs}
The majority of contemporary LLM-based agents, specifically designed to address graph-related tasks, are predominantly tailored for single graph tasks. These agents typically adhere to a one-time-run procedure, aiming to resolve the provided question in a single attempt. Consequently, these agents are neither equipped to function as multi-run interactive agents, capable of adjusting their generated plans based on feedback or additional information, nor are they designed to be user-friendly agents that can effectively manage a wide array of user-given questions. An LLM-based agent~\cite{wang2024survey} that embodies the ideal qualities should not only be user-friendly but also possess the capability to dynamically search for answers within graph data in response to a diverse range of open-ended questions posed by users. This would necessitate the development of an agent that is both adaptable and robust, able to engage in iterative interactions with users and adept at navigating the complexities of graph data to provide accurate and relevant answers.
\section{Conclusion}
\label{sec:conclusion}
In this comprehensive survey, we delve into the current state of large language models specifically tailored for graph data, proposing an innovative taxonomy grounded in the distinctive designs of their inference frameworks. We meticulously categorize these models into four unique framework designs, each characterized by its own set of advantages and limitations. Additionally, we provide a detailed discussion on these characteristics, enriching our analysis with insights into potential challenges and opportunities within this field. Our survey not only serves as a critical resource for researchers keen on exploring and leveraging large language models for graph-related tasks but also aims to inspire and guide future research endeavors in this evolving domain. Through this work, we hope to foster a deeper understanding and stimulate further innovation in the integration of LLMs with graphs.

\clearpage
\bibliographystyle{ACM-Reference-Format}
\balance
\bibliography{sample-base}

\section{Appendix}

In Table~\ref{tab:summary}, we provide an overview of notable graph learning techniques that utilize large language models.

\begin{table*}[h]
    \centering
    \footnotesize
    \caption{Summary of representative graph learning methods with large language models.}
    \resizebox{\textwidth}{!}{
    \begin{tabular}{cccccc}
    \toprule	
    \textbf{Category} & \textbf{Method} & \textbf{Pipeline} & \textbf{Domain} & \textbf{Venue} & \textbf{Year}\\
    \midrule
    \multirow{12}{*}{\rotatebox[origin=c]{0}{GNNs as Prefix}} 
    & GraphGPT~\cite{GraphGPT}                  & Node-level Tokenization  & General Graph        & SIGIR               & 2024 \\
    & HiGPT~\cite{HiGPT}                        & Node-level Tokenization  & Heterogeneous Graph  & KDD                 & 2024 \\
    & GraphTranslator~\cite{GraphTranslator}    & Node-level Tokenization  & General Graph        & WWW                 & 2024 \\
    & UniGraph~\cite{UniGraph}                  & Node-level Tokenization  & General Graph        & arXiv               & 2024 \\
    & GIMLET~\cite{GIMLET}                      & Node-level Tokenization  & Bioinformatics       & NeurIPS             & 2024 \\
    & XRec~\cite{XRec}                          & Node-level Tokenization  & Recommendation       & arXiv               & 2024 \\
    & GraphLLM~\cite{GraphLLM}                  & Graph-level Tokenization & Graph Reasoning      & arXiv               & 2023 \\
    & GIT-Mol~\cite{GIT-Mol}                    & Graph-level Tokenization & Bioinformatics       & Comput Biol Med     & 2024 \\
    & MolCA~\cite{MolCA}                        & Graph-level Tokenization & Bioinformatics       & EMNLP               & 2023 \\
    & InstructMol~\cite{InstructMol}            & Graph-level Tokenization & Bioinformatics       & arXiv               & 2023 \\
    & G-Retriever~\cite{G-Retriever}            & Graph-level Tokenization & Graph-based QA       & arXiv               & 2024 \\
    & GNP~\cite{GNP}                            & Graph-level Tokenization & Graph-based QA       & AAAI                & 2024 \\
    \midrule
    \multirow{10}{*}{\rotatebox[origin=c]{0}{LLMs as Prefix}} 
    & G-Prompt~\cite{G-Prompt}                  & Embs. from LLMs for GNNs  & General Graph        & arXiv               & 2023 \\
    & SimTeG~\cite{SimTeG}                      & Embs. from LLMs for GNNs  & General Graph        & arXiv               & 2023 \\
    & GALM~\cite{GALM}                          & Embs. from LLMs for GNNs  & General Graph        & KDD                 & 2023 \\
    & OFA~\cite{OFA}                            & Embs. from LLMs for GNNs  & General Graph        & ICLR                & 2024 \\
    & TAPE~\cite{TAPE}                          & Embs. from LLMs for GNNs  & General Graph        & ICLR                & 2024 \\
    & LLMRec~\cite{LLMRec}                      & Embs. from LLMs for GNNs  & Recommendation       & WSDM                & 2024 \\
    & OpenGraph~\cite{OpenGraph}                & Labels from LLMs for GNNs & General Graph        & arXiv               & 2024 \\
    & LLM-GNN~\cite{LLM-GNN}                    & Labels from LLMs for GNNs & General Graph        & ICLR                & 2024 \\
    & GraphEdit~\cite{GraphEdit}                & Labels from LLMs for GNNs & General Graph        & arXiv               & 2023 \\
    & RLMRec~\cite{RLMRec}                      & Labels from LLMs for GNNs & Recommendation       & WWW                 & 2024 \\
    \midrule
    \multirow{16}{*}{\rotatebox[origin=c]{0}{LLMs-Graphs Interaction}} 
    & MoMu~\cite{GraphGPT}                      & Alignment between GNNs and LLMs  & Bioinformatics       & arXiv               & 2022 \\
    & ConGraT~\cite{HiGPT}                      & Alignment between GNNs and LLMs  & General Graph        & arXiv               & 2023 \\
    & G2P2~\cite{GraphTranslator}               & Alignment between GNNs and LLMs  & General Graph        & SIGIR               & 2023 \\
    & GRENADE~\cite{UniGraph}                   & Alignment between GNNs and LLMs  & General Graph        & EMNLP               & 2023 \\
    & MoleculeSTM~\cite{GIMLET}                 & Alignment between GNNs and LLMs  & Bioinformatics       & Nature MI           & 2023 \\
    & THLM~\cite{XRec}                          & Alignment between GNNs and LLMs  & Heterogeneous Graph  & EMNLP               & 2023 \\
    & GLEM~\cite{GraphLLM}                      & Alignment between GNNs and LLMs  & General Graph        & ICLR                & 2023 \\
    & GreaseLM~\cite{GreaseLM}                  & Fusion Training of GNNs and LLMs & Graph-based QA       & ICLR                & 2022 \\
    & DGTL~\cite{DGTL}                          & Fusion Training of GNNs and LLMs & General Graph        & arXiv               & 2023 \\
    & ENGINE~\cite{ENGINE}                      & Fusion Training of GNNs and LLMs & General Graph        & arXiv               & 2024 \\
    & GraphAdapter~\cite{GraphAdapter}          & Fusion Training of GNNs and LLMs & General Graph        & WWW                 & 2024 \\
    & Pangu~\cite{Pangu}                        & LLMs Agent for Graphs            & Graph-based QA       & ACL                 & 2023 \\
    & Graph Agent~\cite{GraphAgent}             & LLMs Agent for Graphs            & General Graph        & arXiv               & 2023 \\
    & FUXI~\cite{FUXI}                          & LLMs Agent for Graphs            & Graph-based QA       & arXiv               & 2024 \\
    & Readi~\cite{Readi}                        & LLMs Agent for Graphs            & Graph-based QA       & arXiv               & 2024 \\
    & RoG~\cite{RoG}                            & LLMs Agent for Graphs            & Graph-based QA       & ICLR                & 2024 \\
    \midrule
    \multirow{17}{*}{\rotatebox[origin=c]{0}{LLMs-Only}} 
    & NLGraph~\cite{NLGraph}                      & Tuning-free  & Graph Reasoning       & NeurIPS             & 2024 \\
    & GPT4Graph~\cite{GPT4Graph}                  & Tuning-free  & Graph Reasoning \& QA & arXiv               & 2023 \\
    & Beyond Text~\cite{BeyondText}               & Tuning-free  & General Graph         & arXiv               & 2023 \\
    & Graph-LLM~\cite{Graph-LLM}                  & Tuning-free  & General Graph         & KDD Exp. News.      & 2023 \\
    & GraphText~\cite{GraphText}                  & Tuning-free  & General Graph         & arXiv               & 2023 \\
    & Talk like a Graph~\cite{Talklikeagraph}     & Tuning-free  & Graph Reasoning       & arXiv               & 2023 \\
    & LLM4DyG~\cite{LLM4DyG}                      & Tuning-free  & Dynamic Graph         & arXiv               & 2023 \\
    & GraphTMI~\cite{GraphTMI}                    & Tuning-free  & General Graph         & arXiv               & 2023 \\
    & Ai et al.~\cite{GraphMM}                    & Tuning-free  & Multi-modal Graph     & arXiv               & 2023 \\
    & InstructGLM~\cite{InstructGLM}              & Tuning-required & General Graph        & EACL                & 2024 \\
    & WalkLM~\cite{WalkLM}                        & Tuning-required & General Graph        & NeurIPS             & 2024 \\
    & LLaGA~\cite{LLaGA}                          & Tuning-required & General Graph        & ICML                & 2024 \\
    & InstructGraph~\cite{InstructGraph}          & Tuning-required & General Graph \& QA \& Reasoning        & arXiv               & 2024 \\
    & ZeroG~\cite{ZeroG}                          & Tuning-required & General Graph        & arXiv               & 2024 \\
    & GraphWiz~\cite{GraphWiz}                    & Tuning-required & Graph Reasoning      & arXiv               & 2024 \\
    & GraphInstruct~\cite{GraphInstruct}          & Tuning-required & Graph Reasoning \& Generation      & arXiv               & 2024 \\
    & MuseGraph~\cite{MuseGraph}                  & Tuning-required & General Graph        & arXiv               & 2024 \\
    \bottomrule
    \end{tabular}
    }
    \label{tab:summary}
\end{table*}

\end{document}